\documentclass[10pt, a4paper]{article}
\usepackage{lrec2022} % this is the new LREC2022 Style
\usepackage{multibib}
\newcites{languageresource}{Language Resources}
\usepackage{graphicx}
\usepackage{tabularx}
\usepackage{soul}
\usepackage{amsmath}
\usepackage{titlesec}
\titleformat{\section}{\normalfont\large\bfseries\center}{\thesection.}{1em}{}
\titleformat{\subsection}{\normalfont\SmallTitleFont\bfseries\raggedright}{\thesubsection.}{1em}{}
\titleformat{\subsubsection}{\normalfont\normalsize\bfseries\raggedright}{\thesubsubsection.}{1em}{}
\renewcommand\thesection{\arabic{section}}
\renewcommand\thesubsection{\thesection.\arabic{subsection}}
\renewcommand\thesubsubsection{\thesubsection.\arabic{subsubsection}}
\usepackage{epstopdf}
\usepackage[utf8]{inputenc}

\usepackage[hyphens]{url}
\usepackage[hyperindex,breaklinks]{hyperref}
\hypersetup{
           breaklinks=true,   % splits links across lines
           colorlinks=false,   % displays links as colored text instead of blocks
           pdfusetitle=true,  % \title and \author values into pdf metadata
        }
\usepackage[hyphenbreaks]{breakurl}
        
\usepackage{xstring}

\usepackage{color}

\title{The Norwegian Parliamentary Speech Corpus}

\name{Per Erik Solberg, Pablo Ortiz} 

\address{National Library of Norway, Telenor Research \\
         Oslo, Norway \\
         per.solberg@nb.no, pablo.ortiz@telenor.com \\}

\abstract{
The Norwegian Parliamentary Speech Corpus (NPSC) is a speech dataset with recordings of meetings from Stortinget, the Norwegian parliament.
It is the first, publicly available dataset containing unscripted, Norwegian speech designed for training of automatic speech recognition (ASR) systems.
The recordings are manually transcribed and annotated with language codes and speakers, and there are detailed metadata about the speakers. The transcriptions exist in both normalized and non-normalized form, and non-standardized words are explicitly marked and annotated with standardized equivalents. To test the usefulness of this dataset, we have compared an ASR system trained on the NPSC with a baseline system trained on only manuscript-read speech.
These systems were tested on an independent dataset containing spontaneous, dialectal speech.
The NPSC-trained system performed significantly better, with a {22.9}\% relative improvement in word error rate (WER). Moreover, training on the NPSC is shown to have a ``democratizing'' effect in terms of dialects, as improvements are generally larger for dialects with higher WER from the baseline system.
 \\ \newline  }
\begin{document}

\maketitleabstract

\section{Introduction}
The Norwegian Parliamentary Speech Corpus (\emph{NPSC}) is an open dataset intended for acoustic modelling of Norwegian unscripted speech \citelanguageresource{npsc}. It is developed and distributed by the Language Bank at the National Library of Norway. The dataset consists of about 140 hours\footnote{Including breaks. The speech amounts to about 126 hours.} of recordings of meetings at \emph{Stortinget}, the Norwegian parliament, in 2017 and 2018 with orthographic transcriptions in Norwegian \emph{Bokmål} and Norwegian \emph{Nynorsk}, the two official written standards of the Norwegian language. The dataset is public domain (CC0), and there are, consequently, no restrictions on its use.

While there exist some open, datasets with manuscript-read speech for Norwegian Bokmål, there are few, unscripted datasets suited for acoustic modelling of Norwegian. There is a lack of available speech data of both kinds for Norwegian Nynorsk. The NPSC is intended to fill this gap.

In the remainder of this section, we show why a dataset like the NPSC is needed and list some existing datasets for Norwegian ASR. In section \ref{sec:npsc}, we explain what the NPSC dataset contains. Section \ref{sec:makingnpsc} lays out how the NPSC was developed. Section \ref{sec:eval} reports on an ASR experiment where we have used the NPSC for training and testing. Finally, section \ref{sec:discussion} raises some points for discussion and suggests some avenues for further development, and section \ref{sec:conclusion} concludes the paper.

\subsection{Why Norwegian ASR is challenging}
\label{challenging}
Norwegian is the native language of most of Norway's 5.3 million inhabitants. Even though the linguistic community is relatively small, the language is quite diverse, which makes ASR particularly challenging. Firstly, as mentioned, there are two official written standards. Secondly, Norwegian has many dialects, which differ lexically, grammatically and pho\-no\-lo\-gi\-cal\-ly. There is no spoken standard of Norwegian, and speakers use dialects even in official settings \cite{roynelanddialects}. It is likely that many speakers also use their dialect, or would prefer to use it, when speaking with speech assistants, smart-home devices, dictation software and other kinds of technology with a voice user interface. High quality datasets for acoustic modelling of Norwegian therefore require speech data in different dialects, and should include transcriptions in both written standards.
Finally, the Bokmål and Nynorsk standards allow for a lot of options: many words have multiple alternative spellings or inflectional variants, which usually correspond to dialectal variation in the spoken language. When testing an ASR system, the predicted transcription and the gold-standard transcriptions may contain different variants of the same word, e.g. \textit{vet} and \textit{veit}, `know'. This will be counted as an error, but will not render the transcription less intelligible or be perceived as a grave error by users.

\subsection{Existing Datasets for Norwegian ASR}
\label{sec:data}
The Language Bank at the National Library of Norway is the largest provider of open-source datasets for Norwegian speech technology. There are several speech datasets in the Language Bank. The largest is the ASR dataset made by the defunct firm Nordisk språkteknologi (\emph{NST}) at the turn of the millennium \citelanguageresource{nst}. This dataset consists of 540 hours of recordings of close to 1000 informants reading from manuscripts. The manuscripts contain mostly sentences, but also sequences of numbers and repeated words. The corpus also includes a written version of the manuscript sentences and metadata about the speakers (age, gender, region of birth and region of youth). While the NST dataset is well suited for fundamental ASR of Norwegian, it has some limitations. Being a manuscript-read dataset, it only contains planned speech, and consequently provides less evidence of hesitations, interruptions and other speech phenomena which are common in unscripted speech. Since the speakers read sentences in Bokmål, the dataset does not contain, or contains to a very limited degree, dialectal phenomena which deviate from the Bokmål norm.
ASR systems typically perform less well when applied to a dialect they have not been exposed to during training \cite{elfeky2016towards}. To train a general purpose ASR system which handles dialects well, it would be advantageous to supplement this dataset with transcribed recordings of unscripted speech of speakers from various parts of the country. Also, there are no Nynorsk transcriptions in the dataset.

Prior to the publication of the NPSC, the Language Bank distributed one dataset with spontaneous speech: Module 3 of the NB Tale dataset \citelanguageresource{nbt}.
This module contains transcribed recordings of 365 speakers, native and non-native, speaking for 2 minutes on a subject of their choice. The recordings of the 229 speakers with Norwegian as their native language amounts to between 7 and 8 hours. While this dataset is rather small, it is valuable for testing the performance of ASR systems on dialects, as the speakers are divided into fine-grained dialect groups.\footnote{The datasets mentioned here, as well as smaller speech datasets for speech synthesis and dictation, are found in the repository of the Language Bank:  \url{https://www.nb.no/sprakbanken/en/resource-catalogue/?_type=speech&_origin=language-bank}}

Finally, the University of Oslo has made many of their dialect corpora available for download.\footnote{\url{http://tekstlab.uio.no/LIA/filer/}} These corpora are not developed with speech technology in mind, and to our knowledge, they have not yet been used for ASR development and testing. They could, however, be an interesting source of data.

The aim of the NPSC project was to supplement the existing resources with a unscripted dataset for training and testing ASR systems.

\section{The Content of the NPSC}
\label{sec:npsc}
\subsection{The Audio Files}
The NPSC consists of recordings of entire days of parliamentary debates from 2017 and 2018, 41 in total.\footnote{If the debate lasts for more than 6 hours, the recording in the NPSC is cut at about 6 hours and 10 minutes.} The length of these recordings varies from less than an hour to 6 hours and 10 minutes. In addition to the audio files of the entire meetings, there are also segmented audio files for each sentence in the corpus. The audio files are in the wav format with two channels, a sampling rate of 48 kHz, and a bit depth of 16 bits.\footnote{Note, however, that the audio files are extracted from Stortinget video files, which are compressed. We have not been able to obtain uncompressed audio files.}

\subsection{The Transcriptions}
The audio files are transcribed sentence by sentence. Each sentence is annotated with a manually specified start and end time, as well as the name and identifier of the speaker, and is transcribed in Norwegian Bokmål or Norwegian Nynorsk. Every speaker in the corpus is transcribed consistently in one written standard (unless the speaker is quoting something in the other standard). We have not chosen the written standard on an independent basis, but follow the official proceedings from Stortinget, which, in turn, use the written standard each speaker prefers. This gives a percentage of Norwegian Nynorsk of about 13\%.

The transcriptions exist in different versions:
\begin{enumerate}
    \item A sentence-segmented, non-normalized version. In this version, numbers, dates and years are written with letters instead of digits, and abbreviations are not used. This is probably the most suited version for acoustic modelling, as the transcriptions are the most faithful to the pronunciation. 
    \item A sentence-segmented, normalized version. Here, numbers, dates and years are written with digits in standardized formats, and common abbreviations are used. This version is generated from the non-normalized version via normalization rules, which are provided with the corpus for reference. Both the normalized and non-normalized transcriptions contain dialect words and other, non-standard words. Information about standardization is found in the word-tokenized versions of the corpus.
    \item A sentence-segmented, normalized version where Bokmål transcriptions are machine-translated to Nynorsk and Nynorsk transcriptions are machine-translated to Bokmål using the open-source translation system Apertium \cite{forcada2011apertium}. This version is probably not suited for acoustic modelling, but might be useful, e.g., for language modelling of Nynorsk.
    \item A word-tokenized, non-normalized version. In this version, each word contains metadata about whether or not it is standardized. If the word is not standardized, an equivalent, standardized word is given in a separate field. There is also metadata indicating if a word is interrupted.
    \item A word-tokenized, normalized version with the same metadata as in the above word-tokenized version, but with normalized numbers, dates, years and abbreviations.
\end{enumerate}

A list of the speakers is also included with the NPSC with metadata about their name, gender, date of birth, place of birth, region of birth, electoral district, dialect, written standard and Wikidata URI.

Table \ref{tab:corpstats} lists some corpus statistics.

\begin{table}[!h]
\begin{center}
\begin{tabularx}{\columnwidth}{|l|X|}
      \hline
      Duration, pauses included & 140.3 hours \\
    \hline
      Duration, pauses excluded & 125.7 hours \\
      \hline
      Word count & 1.2 million\\
      \hline
      Sentence count & 64 531\\
      \hline
      Language distribution & Nynorsk: 12.8\%\\  & Bokmål: 87.2\%\\
      \hline
     Gender distribution & F: 38.3\%, M: 61.7\%\\
      \hline

\end{tabularx}
\caption{Corpus statistics.}
 \label{tab:corpstats}
 \end{center}
\end{table}

\section{Making the NPSC}
\label{sec:makingnpsc}
\subsection{Choice of Texts}
There are several advantages to using Stortinget data for an open speech dataset for acoustic modelling. Firstly, the data are public domain, and can, therefore, be reshared without restrictions, unlike, e.g., broadcast audio, where copyright and privacy concerns makes resharing challenging. Secondly, the speakers are public figures, and we have access to detailed metadata about them from public sources. Thirdly, there are official proceedings of the Stortinget meetings. These are not verbatim transcriptions, but they render what is said in the meetings quite faithfully. They can, therefore, be used in the preprocessing of the transcriptions (see below). Finally, the representatives come from all over the country and tend to use their dialect, so there is a good dialect distribution.

The Stortinget data have some disadvantages too. Some of what is said in the meetings is read from a manuscript, so the corpus does not consist entirely of unplanned speech. Furthermore, parliamentary meetings have a particular style and vocabulary which may differ from other domains. We have attempted to compensate for this, at least to some degree, by working with the Stortinget stenographers to identify meetings with a high amount of unplanned speech and varied vocabulary.

\subsection{Preprocessing and Transcription}
Prior to manual transcription of a Stortinget meeting, the audio file was run through Google Cloud Speech-To-Text.\footnote{\url{https://cloud.google.com/speech-to-text}} A Python script compared the ASR transcription with the official proceedings from Stortinget and replaced words from the transcription with words at the same location in the proceedings with a short edit distance from the ASR word. This improved the automatic transcription noticeably. It also added Nynorsk words at appropriate places, despite the fact that the Google ASR only produces Bokmål. Transcribers then corrected the automatic transcriptions in a tailor-made, web-based transcription tool. After transcription, another transcriber reviewed the transcription and corrected errors.

The transcribers were all trained linguists or philologists. The transcription guidelines were written by the core team of transcribers during the first phase of the project \cite{npscguidelines}. The guidelines set up detailed procedures for handling dialect words and other non-standard words. Whenever transcribers encountered such words, they wrote both the dialect word and a standardized equivalent, which can be found in the word-tokenized version of the transcriptions. They also maintained word lists of such instances so that non-standard words were transcribed as consistently as possible.

\subsection{Postprocessing and Dialect Annotation}
When all the meetings were transcribed, the transcripts were run through a correction script that corrected common errors. Furthermore, they were processed with normalization grammars that produced the normalized version of the transcriptions, as well as a machine translation pipeline that produced the translated version.

We used the speaker names, added by the transcribers, to run queries with the Wikidata SPARQL endpoint\footnote{\url{https://query.wikidata.org/}} and extracted metadata about the speakers. A linguist on the team listened to the longest sentence of each speaker and determined which region (East, West, South, Trøndelag, North or unknown) their dialect came from. This dialect classification is quite coarse-grained. However, if users couple it with the metadata about place of birth, it is possible to make more fine-grained assumptions, at least when the dialect region and the region of birth match.

\subsection{Data Splits}
The dataset is split into a training, evaluation and test set. We did not make a random selection of sentences for each split, as is often done. Instead, entire meetings were selected for each split. The motivation for splitting the data in this way was to make it possible to train and test systems that use context beyond the sentence, such as \cite{ortiz2021disambiguationbert}, and to minimize the overlap of topics, speakers and vocabulary across the splits, such that testing is more realistic. We made an effort to get a similar distribution of Bokmål and Nynorsk and male and female speakers in each split, and we also checked that each dialect region was reasonably represented across the splits. We tried to stay as close to a 80-10-10 percent split as possible. There are 51278, 6838 and 6344 sentences in the training, evaluation and test splits respectively.

\section{Evaluating the Dataset}
\label{sec:eval}

In this section we perform a set of ASR experiments with two main purposes:
\begin{itemize}
    \item Benchmark Norwegian ASR models on ``clean'' data such as NST, and on more realistic data such as NPSC. With it we want to emphasize the need for more data of the same kind as NPSC.
    \item Measure the relative improvement of ASR models after adding NPSC data when testing on spontaneous, dialectal speech such as that in NB Tale. 
\end{itemize}
In the following subsections we describe our ASR system and the models used in the experiments. Then we present the results obtained for different models, datasets and dialects.

\subsection{Baseline ASR System}
\label{subsec:baseline}
Our baseline ASR system is based on Deepspeech 2 \cite{amodei2016deep}, where the acoustic model (AM) is combined with an n-gram language model (LM) during decoding. All the LMs described below are trained using the implementation of Kneser-Ney smoothed n-gram estimation from \cite{heafield_scalable_2013}\footnote{\url{https://github.com/kpu/kenlm}}. We refer the reader to \cite[Section II]{ortiz2021disambiguationbert} for a detailed description of the architecture and code base for the baseline model.

We train the primary AM with a refined\footnote{We removed utterances with less than three words and those containing only three repetitions of the same word, since those are more appropriate for dictation.} version of NST data consisting of 394.5 hours, where 300h are used as the training dataset. For the LM we use a  5-gram model trained with approximately 13 million sentences from a non-public corpus gathered by NST consisting of newspaper text, denoted by LM$_\text{base}$.\footnote{We thank August Moum and Skjalg Winnerdal for providing access to this model. It was trained with a version of the newspaper corpus curated at the Norwegian University of Science and Technology (NTNU) during the SVoG project in collaboration with other institutions.}

\subsection{Models with NPSC Data}
\label{subsec:npscmodels}

We build an acoustic model including NPSC data by fine-tuning the primary AM pre-trained with NST data described above. That is, we take the primary AM as a starting point and train it only on NPSC data. We denote this acoustic model by NPSC$_\text{NST}$. 

As for the language models, in addition to LM$_\text{base}$, we also built a 3-gram model with NPSC training data, LM$_\text{NPSC}$,  and another 3-gram model with NPSC training data and the transcripts of the NST data used for the acoustic model, LM$_\text{NST+NPSC}$. We use those to benchmark performance on NPSC test data. Still, our main objective is to test how the NPSC data aids when transcribing spontaneous, dialectal speech from an independent dataset, namely Module 3 of NB Tale. 

\subsection{Experiments and Results}
\label{subsec:results}

We test different combinations of the acoustic and language models described above. For that we use three different datasets: the test split of NPSC data, the full Module 3 of NB Tale divided in sentences, and a version of the NST test split which only contains long, fully grammatical sentences, denoted by NST'.

We remark that the objective is not to optimize the absolute performance on NB Tale. In that case, one would fine-tune the AM on a NB Tale training set, build specific LMs, and use more elaborated methods that include speech context \cite{ortiz2021disambiguationbert}, as the ASR models discussed here consider every utterance independently of the context.

For each experiment, we optimize the weights of language model and word count terms on the evaluation split using Optuna \cite{akiba_optuna_2019}. We then use those parameters during test with a beam size of 512 to calculate the average word error rate (WER) per utterance, which gives the results reported in Table \ref{tab:results}.

\begin{table}[h!]
\centering
\setlength{\tabcolsep}{3.1pt}
\begin{tabular}{|c|c|c|c|c|c|}
\hline
Name & {AM} & {LM} & {NST'} & {NPSC} & {NB} \\
\hline
M$_1$ & NST & LM$_\text{base}$ & 2.9 & 40.6 & 48.4 \\
M$_2$ & NPSC$_\text{NST}$ & LM$_\text{NST+NPSC}$ & - & 15.9 & 39.6 \\
M$_3$ & NPSC$_\text{NST}$ & LM$_\text{base}$ & - & 17.8 & 37.3 \\
M$_4$ & NPSC$_\text{NST}$ & LM$_\text{NPSC}$ & - & 17.1 & - \\
\hline
\end{tabular}
\caption{WER(\%) obtained with different combinations of models on our test sets NST', NPSC and Module 3 of NB Tale (NB).}
\label{tab:results}
\end{table}

Our results show that, while our primary model M$_1$ is suited for clean data (NST'), performance on more realistic datasets is severely damaged. By fine-tuning the primary AM with NPSC data (M$_3$), performance on NB Tale is improved by an absolute 11.1\% in WER, or a relative 22.9\%. As expected, performance during testing on NPSC is greatly boosted when fine-tuning the primary AM on the same kind of data. Moreover, as argued above, further improvements on NPSC are observed when applying smaller but more specialized LMs specific to NPSC data as in models M$_2$ and M$_4$. For the sake of simplicity, we do not test M$_4$ on NB Tale data, as the results will be qualitatively the same as those from M$_2$, only with slightly worse performance due to the smaller language model.

Last, for each of the three tests on NB Tale, we evaluate performance on each of the 12 dialect groups featured in the dataset \cite{nbtaledialects}, as shown in Table \ref{tab:dia_results}.

\begin{table}[h!]
\centering
\setlength{\tabcolsep}{5.5pt}
\begin{tabular}{|l|c|c|c|c|}
\hline
\textbf{Dialect area} & {N$_\text{utt}$} & {M$_1$} &
{M$_2$} & {M$_3$} \\
\hline
Agder & 367 & 44.9 & 36.2 & 33.4  \\
Brønnøysund & 440 & 52.6 & 42.2 & 40.0  \\
Finnmark & 314 & 42.9 & 36.1 & 34.2  \\
Møre og Romsdal & 409 & 52.5 & 41.2 & 39.1  \\
Nordland & 311 & 49.1 & 39.7 & 38.0  \\
Oslo & 394 & 39.0 & 33.9 & 31.3  \\
Rogaland-Bergen & 382 & 48.5 & 41.8 & 39.0  \\
Rogaland-Hordaland & 392 & 50.2 & 44.1 & 41.6  \\
Sogn og Fjordane & 396 & 55.0 & 43.1 & 41.0  \\
Troms & 374 & 47.1 & 38.0 & 36.3  \\
Trøndelag & 383 & 55.8 & 42.9 & 41.4  \\
Østlandet & 420 & 42.3 & 35.1 & 32.6  \\
\hline
Weighted average & 382 & 48.4 & 39.6 & 37.3  \\
Standard deviation & 38.1 & 5.28 & 3.54 & 3.67  \\
Rel. st. dev. (\%) & 9.97 & 10.89 & 8.94 & 9.81  \\
\hline
\end{tabular}
\caption{WER(\%) per dialect area in Module 3 of NB Tale for the three combinations of models tested on these data, see Table \ref{tab:results}. The number of utterances per dialect area, N$_\text{utt}$, is used for the weighted average, and the relative standard deviation is simply the standard deviation normalized by the average.}
\label{tab:dia_results}
\end{table}

The results in Table \ref{tab:dia_results} show that after fine-tuning the acoustic model with NPSC data (M$_2$, M$_3$), performance improves substantially across all dialect groups, both when the language model is specific (M$_2$) and when it is left unchanged (M$_3$). Moreover, the relative improvement is generally larger for the dialects with higher WER under the primary model without NPSC data (M$_1$). This means that the NPSC data has a ``democratizing'' effect in terms of dialects. Another way to see this is by analyzing the relative standard deviation of the WER across dialect groups. While the variation across dialects with M$_1$ (10.89\%) is larger than expected given the variation in sample sizes N$_\text{utt}$ (9.97\%), models with NPSC data reduce this variation by a relative 17.94\% and 9.92\%, respectively for M$_2$ and M$_3$.

\section{Discussion}
\label{sec:discussion}
The NPSC models in the experiments reported above are trained on a mix of Bokmål and Nynorsk transcriptions and hence produce a mix of the two written standards. With this setup, a proportion of Nynorsk of about 13\% is reasonable, as it is about the same as it is used in the population at large \cite{ssb2020}. In other words, the models reflect the actual usage in the population. However, a system that produces mixed transcriptions is not desirable in many real-life use-cases, where a transcription system is expected to produce one or the other written standard consistently.

In both the NPSC and NB Tale, non-standard forms of words are explicitly marked, and an alternative, standardized form is given. In the experiments reported here, transcriptions with non-standard vocabulary are used in the training and test data, and the standardized equivalents have been ignored. Consequently, the system produces non-standard words. However, since the NPSC provides extensive metadata on non-standard forms, it is a valuable and useful resource for investigating the mixture of spoken and written forms in ASR.

A different, but related issue is the treatment of standardized variants of the same word during testing. The WER metric qualifies such equivalent written forms of the same word as a full error, e.g. $\text{WER}(\text{vet}, \text{veit})=1$ (cf. section \ref{challenging}). In cases like this, human perception of transcription quality fully disagrees with WER. Softer measures using word embeddings \cite{Le2016BetterEO} can alleviate this discrepancy. However, a model trained on mixed transcriptions and less penalized for mixing equivalent forms would produce a higher mixture of these. We think this topic deserves further investigation as well.

Last, we note that fillers and hesitations are present and explicitly marked in NPSC. These events do not appear in other datasets with manuscript-read speech, which makes NPSC a useful resource for the study of such acoustic events more typical of spontaneous speech.

\section{Conclusion}
\label{sec:conclusion}
In this paper we have presented the NPSC, an open speech dataset intended to improve ASR for Norwegian spontaneous speech and dialects. In our experiments, the NPSC-trained system performed significantly better than the baseline when tested on Module 3 of NB Tale, with a relative improvement of 22.9\%. Moreover, training on the NPSC has a beneficial effect on the recognition of dialects. There was not only a substantial improvement across all dialects compared to the baseline, but also the improvements were larger for dialects with higher WER in the baseline results, i.e. the relative difference across dialect groups was reduced.

The NPSC is an important contribution to the Norwegian ASR community, as it provides excellent training material with notable variability in spoken and written Norwegian. As such, it enables further research, development and application of ASR not only for Norwegian, but also for many other languages affected by the phenomena we discussed here. Nevertheless, more open data of this kind would be beneficial to keep bringing the applicability of low-resource ASR closer to realistic situations.

\section{Acknowledgements}

We are grateful for useful discussions with Torbjørn Svendsen and Knut Kvale. Thanks also to Andrea Myklebust Huus, Håvard Østli, Marie Røsok and all the others who contributed to the NPSC project. This work has been partially supported by the Norwegian Research Council through the IKTPLUSS grant for the SCRIBE project\footnote{\url{https://scribe-project.github.io/}} (KSP21PD). 

% \nocite{*}
\section{Bibliographical References}\label{reference}
%\label{main:ref}

\bibliographystyle{lrec2022-bib}
\bibliography{NPSC_paper}

\section{Language Resource References}
\label{lr:ref}
\bibliographystylelanguageresource{lrec2022-bib}
\bibliographylanguageresource{languageresource}

\end{document}